%% file: acl_latex.tex
\pdfoutput=1

\documentclass[11pt]{article}

\usepackage[preprint]{acl}

\usepackage{times}
\usepackage{latexsym}

\usepackage[T1]{fontenc}

\usepackage[utf8]{inputenc}

\usepackage{microtype}

\usepackage{inconsolata}

\usepackage{graphicx}

\title{StateAct: Enhancing LLM Base Agents via Self-prompting and State-tracking}

\author{Nikolai Rozanov \\
  Imperial College London \\
  Department of Computing \\
  \texttt{nikolai.rozanov13@imperial.ac.uk} \\\And
  Marek Rei \\
  Imperial College London \\
  Department of Computing \\
  \texttt{marek.rei@imperial.ac.uk} 
  }

\usepackage{subcaption} 
\usepackage{listings}
\lstdefinestyle{mystyle}{
    backgroundcolor=\color{lightgray},   
    commentstyle=\color{green},          
    keywordstyle=\color{blue},           
    numberstyle=\tiny\color{gray},       
    stringstyle=\color{red},             
    basicstyle=\ttfamily\footnotesize,   
    breakatwhitespace=false,             
    breaklines=true,                     
    captionpos=b,                        
    keepspaces=true,                     
    showspaces=false,                    
    showstringspaces=false,              
    showtabs=false,                      
    tabsize=2,                           
}
\usepackage{booktabs}
\usepackage{xcolor, colortbl}
\definecolor{lightgreen}{RGB}{200, 255, 200}
\definecolor{darkgreen}{RGB}{100, 200, 100}

\begin{document}
\maketitle

\input{sections/abstract}
\input{sections/1_introduction}

\input{sections/2_background_v2}

\input{sections/3_method}
\input{sections/4_experiments}

\input{sections/5_results}
\input{sections/6_analysis}
\input{sections/7_conclusion}
\input{sections/8_ethics}

\input{sections/9_limitations}

\bibliography{acl_latex}

\input{sections/appendix}

\end{document}

%% file: sections/abstract.tex
\begin{abstract}
Large language models (LLMs) are increasingly used as autonomous agents, tackling tasks from robotics to web navigation. Their performance depends on the underlying \textit{base agent}. Existing methods, however, struggle with long-context reasoning and goal adherence.  
We introduce \textbf{StateAct}, a novel and efficient base agent that enhances decision-making through (1) \textit{self-prompting}, which reinforces task goals at every step, and (2) \textit{chain-of-states}, an extension of chain-of-thought that tracks state information over time.  
StateAct outperforms ReAct, the previous best \textit{base agent}, by over 10\% on Alfworld, 30\% on Textcraft, and 7\% on Webshop across multiple frontier LLMs. We also demonstrate that StateAct can be used as a drop-in replacement for ReAct with advanced LLM agent methods such as test-time scaling, yielding an additional 12\% gain on Textcraft.  
By improving efficiency and long-range reasoning without requiring additional training or retrieval, StateAct provides a scalable foundation for LLM agents. We open source our code to support further research at
\url{https://github.com/ai-nikolai/stateact}.
\end{abstract}

%% file: sections/1_introduction.tex
\section{Introduction}
 Leveraging the in-built world and commonsense knowledge\footnote{Commonsense- and world- knowledge as explored by \citet{lauscher-etal-2020-common}, for example.} of large language models (LLMs), such as GPT, Gemini and Mixtral \cite{brown2020language, Anil2023GeminiAF, jiang2024mixtral} to perform interactive reasoning tasks
has become a frontier in AI research. ``AI Agents'' are now able to solve a range of multi-modal complex tasks \cite{durante2024agent}.

These include simulated robotics tasks \cite{puig2018virtualhome, shridhar2021alfworld} and digital tasks, such as online shopping \cite{yao2023webshop}, navigating operating systems \cite{liu2023agentbench}, and playing a variety of games \cite{côté2019textworld,liu2023agentbench, prasad2024adaptasneededdecompositionplanning}.

At the core of an LLM agent is the \textit{base agent}, such as Act \citep{huang2022language}, ReAct \citep{yao2023react}, and AdaPlanner \citep{Sun2023AdaPlannerAP}. Existing efforts to improve LLM agents build on top of base agents and are usually quite resource-intensive: \citet{Wu2024StateFlowEL} require human expert annotations of rules; 
\citet{Sun2023AdaPlannerAP} require a code execution environment with carefully crafted code-based prompts; 
\citet{Fu2024AutoGuideAG} use additional training data together with retrieval augmented generation (RAG) to help the AI agent; \citet{yang2024react} use additional training data to fine-tune the LLM; \citet{shinn2023reflexion} and \citet{prasad2024adaptasneededdecompositionplanning} use test-time computation to produce better results. Notably, most of the current state-of-the-art methods use ReAct as the base agent.

In our work, we propose a new base agent, called StateAct. Starting with the observation that: i) LLM agents fail to follow the original instruction and goal in longer interactions; and ii) LLMs struggle with long-context despite longer available contexts \citep{Li2023LooGLECL, coelho2024dwell}, we propose two main contributions for improving the base agent. To address i), we propose a mechanism for the agent to `self-prompt' at every turn of the interaction to improve staying on track with the main goal. Our agent `reminds' itself of the goal at every turn. To address ii), we propose `chain-of-states', an extension of chain-of-thought based on state tracking to help the agent stay on track with the current interaction and context. The agent keeps track of its state in the environment (such as location and inventory).

Our experiments show that StateAct achieves near state-of-the-art performance without using additional training data or additional tools, and significantly outperforms ReAct across multiple tasks and eight frontier LLMs of varying sizes. Specifically, StateAct improves performance over ReAct by more than 10\% on Alfworld \citep{shridhar2021alfworld}, 30\% on Textcraft \citep{prasad2024adaptasneededdecompositionplanning}, and 7\% on Webshop \citep{yao2023webshop}.  

Additionally, we validate that StateAct can serve as a drop-in replacement for existing extension methods. Using test-time computation \citep{prasad2024adaptasneededdecompositionplanning}, we achieve a further 12\% performance gain with StateAct on Textcraft using ADaPT.

%% file: sections/2_background_v2.tex
\section{Background}
\begin{figure*}[t]
  \includegraphics[width=1\textwidth]{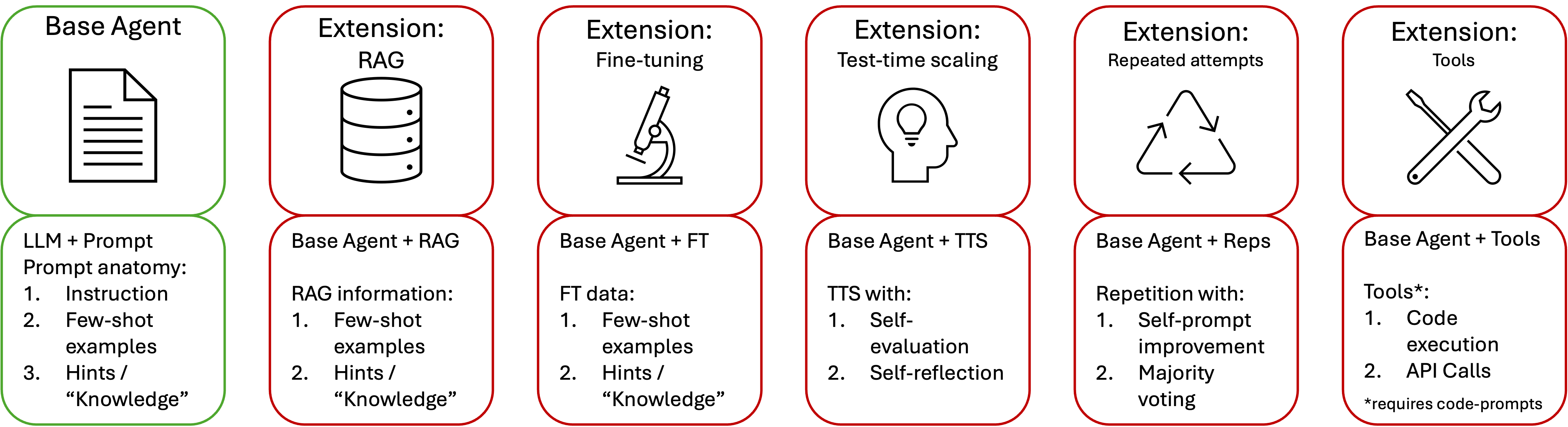}
  \caption{Overview of LLM-agent methods and their components.}
  \label{fig:background}
\end{figure*}
Due to recent advances, LLMs are now being used as autonomous agents in interactive environments as an alternative to traditional reinforcement learning (RL) \cite{Sutton2018, yao2023react, li2022pretrained, Nottingham2023SelectivePO}. LLM agents now tackle tasks in simulated environments such as Alfworld, Webshop, and Textcraft.
LLM agents consist of the \textit{base agent} and a possible extensions, such as fine-tuning (FT), retrieval augmented generation (RAG), test-time scaling (TTS), repeated attempts (REPS) or tools; see Table \ref{fig:background}. \textbf{StateAct is a base agent that can be used in combination with extensions}.

\subsection{Base agents}
\citet{huang2022language, huang2022inner} were among the first to use LLMs directly to act in an interactive environment; their method produces agent actions as output after receiving environment observations as input.
ReAct \cite{yao2023react}  take this work further by combining \textit{acting} \cite{huang2022language} and \textit{chain-of-thought} \cite{wei2023chainofthought}. 
Progprompt and AdaPlanner \citep{singh2022progprompt, Sun2023AdaPlannerAP} use code-based prompts to interact with environments. 
Notably, \textbf{ReAct is the base agent of modern state-of-the-art approaches}. Our method, StateAct, falls in the category of base agents and is therefore most comparable to ReAct.


\subsection{Extensions of base agents: RAG, fine-tuning, test-time-scaling}
ExpeL \cite{zhao2023expel} extend ReAct by using additional training data to generate success trajectories during training. At inference time, they look up the closest success trajectories as few-shot examples to the agent. Follow-on work, AutoGuide \cite{Fu2024AutoGuideAG}, uses ReAct as the base agent with additional training data to create state-aware text-based guidelines. This `knowledge' of the environment is then used with retrieval augmented generation (RAG) to guide the decision-making process. While AutoGuide achieves the best result among RAG approaches, the complexity of the setup makes it less scalable in practice.

\citet{chen2023fireactlanguageagentfinetuning, yao2023react} introduce fine-tuning of ReAct with marginal improvements. KnowAgent \citep{zhu2025knowagentknowledgeaugmentedplanningllmbased} compile knowledge of the environment and distill this into the LLM to produce better results. The best approach, ActRe \cite{yang2024react}, achieves successful fine-tuning of ReAct by annotating successful trajectories with CoT before fine-tuning. Fine-tuning yields the best performance for a specific task, however, requires additional training, which is costly, and does not allow for generalisation. 

ADaPT \citep{prasad2024adaptasneededdecompositionplanning} and THREAD \citep{schroeder2024threadthinkingdeeperrecursive} use test-time computation to achieve better results. This approach is in line with works such as tree-of-thought \citep{Yao2023TreeOT}, where the AI agent proposes several thoughts or actions in one go during test-time inference and another model (often the same LLM) evaluates these; the top k promising thoughts or actions are then expanded upon in a beam-search manner. 
These methods produce strong results and are very easy to setup, albeit they often require more compute budget compared to base agents. In our work we combine ADaPT with StateAct.

\subsection{Alternative methods: tools, hand-crafted rules, multi-agent}
ProgPrompt \citep{singh2022progprompt} and follow-on work AdaPlanner \citep{Sun2023AdaPlannerAP} introduce a code-based prompts \cite{li2023chainofcode}. They use code-execution as an additional tool, by executing LLM-generated code and feeding the results into the next LLM generation. The shortcoming of such code-based prompts is that they require human experts to annotate very long prompts. This can be hard to scale to new environments and requires an additional step of code-execution.

StateFlow \cite{Wu2024StateFlowEL} uses Finite State Machines (FSMs) combined with LLMs to solve Alfworld. These FSMs are human expert-crafted states, transitions and rule-based heuristics, where the LLM is asked to perform limited tasks in each of the given states. While this method can achieve a very high score, it is limited, as it requires a human expert to design the FSM.

Another approach is to use multiple LLMs to `chat' to one another to produce a result. Multi-agent frameworks \citet{wu2023autogen} only achieve minor improvements over using a single agent.

\subsection{State tracking in LLM-based agents}

\citet{chen2024llmstate} propose state-tracking as a way to help the agent solve the task without training data. Their method differs from StateAct two-fold. Firstly, they employ a complex sequence of components working together, which are an LLM-based attention over the observations, an LLM-based compilation of a complex state and a prediction of a program. Secondly, their system involves execution of actual programs. StateAct, on the other hand, requires a straightforward extension of CoT and uses a single LLM call to produce the state, thought and action. Additionally, it does not require program execution. STATLER \citep{yoneda2024statlerstatemaintaininglanguagemodels} also introduce state-tracking for LLM agents. While the state has similarity to StateAct, there are notable differences. Firstly, Statler is aimed at lower level robotics execution and works with domain specific functions. Secondly, Statler produces and requires code to update and read from the state. This complex construction of the state is hard to scale to new environments.


%% file: sections/3_method.tex
\section{Method}

StateAct is an LLM-based AI agent that works on top of pre-trained LLMs. It takes the \textit{textual} `observation' from the environment and, after a single call to the pre-trained LLM, returns the `action' back to the environment, without the use of additional tools or resources.

\begin{figure}[t]
  \includegraphics[width=1\columnwidth]{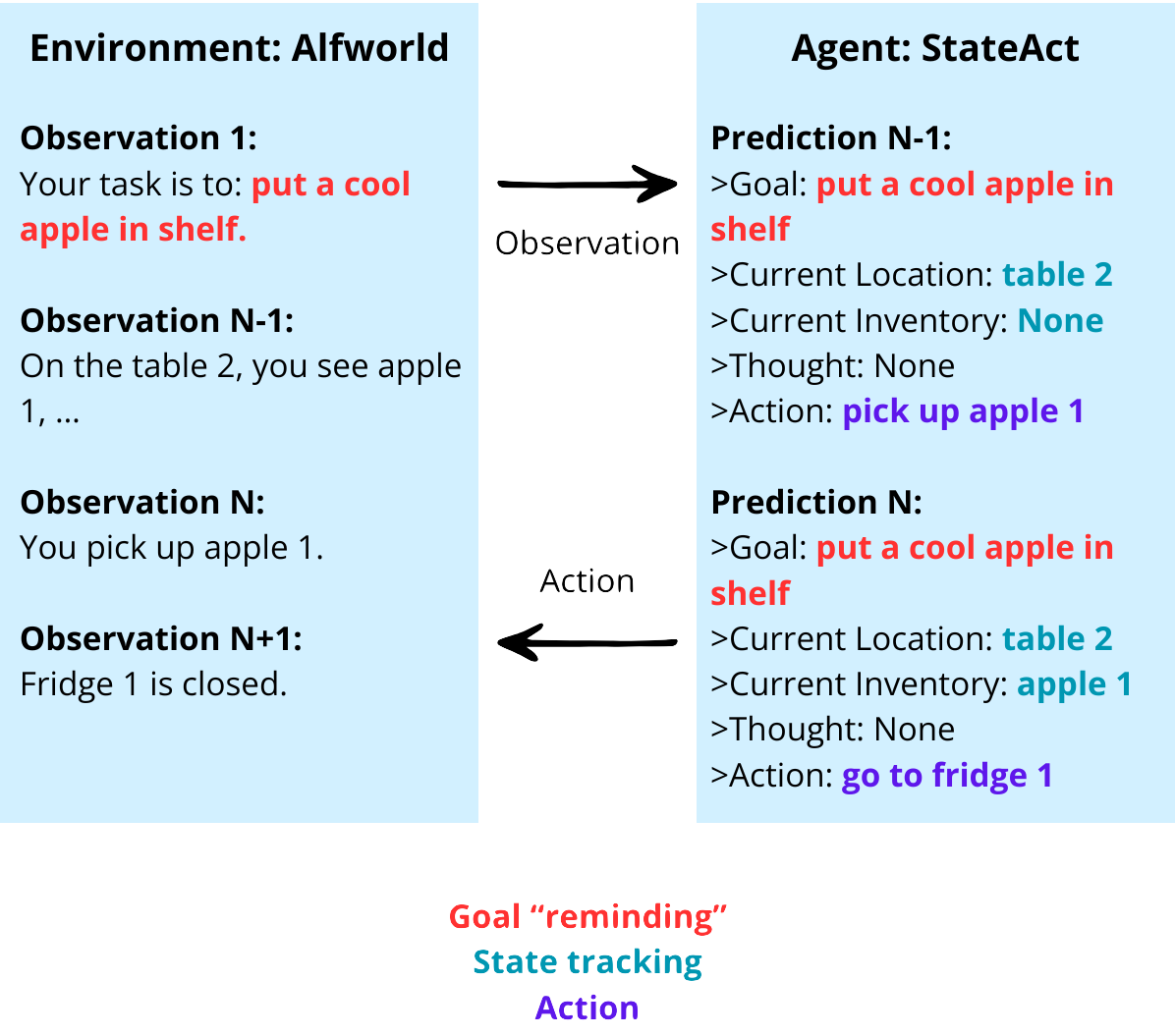}
  \caption{Example of StateAct outputs in Alfworld.}
  \label{fig:stateact_method}
\end{figure}

StateAct utilises in-context learning \cite{brown2020language, wei2023chainofthought} to make the agent interact with the environment. At the core of the approach is a prompt that consists of few-shot examples of successful interaction traces as well as the current interaction trace up to the current step in the environment.
An interaction trace consists of alternating observations from the environment and desired (or actual) outputs from the LLM. In the case of StateAct, the LLM is tasked to generate the `goal', `state', `thought' and `action'. The action is then extracted and passed to the environment to produce the next observation, see Figure \ref{fig:stateact_method}. This renders StateAct similar to ReAct and therefore an easy replacement for extension methods.



\subsection{Self prompting}
One of the key ideas of StateAct to overcome the `haystack' challenge for long prompts and therefore long horizon problems \citep{coelho2024dwell} is to introduce a mechanism for the LLM to pass an instruction to itself. This approach can be applied in many settings and to alleviate various problems (including goal reminding and formatting). In our setting, we focus on keeping the Agent on track with the main goal and so the LLM reminds itself of this goal at every turn.

\subsection{Chain of states as state-tracking}
The idea of state-tracking is to introduce `structured thoughts' into the reasoning part of the LLM Agent, specifically by giving the agent small intermediate predictions that can be inferred from the environment and actions. This method is different to existing methods such as ReAct \citep{yao2023react}, where CoT is taken to mean verbal `thoughts'. The inspiration comes from the original CoT paper \citep{wei2023chainofthought} where the LLM is tasked with producing intermediate calculations. For StateAct, the LLM agent is tasked with predicting very specific intermediate steps, such as the current location or the inventory of the agent. \textbf{See Appendix \ref{sec:appendix_formalizing}  for a formalization of StateAct.}


\subsection{An example trace}
An example of the trace for step \textit{n} in Alfworld is:\\

\textit{Observation n:} \\\texttt{On the countertop 2, you see a tomato 1.}

\textit{LLM Output n:} \\\texttt{>Goal: Put a clean tomato in fridge\\
Current Location: countertop 2\\
Current Inventory: None\\
Locations Visited: table 1, countertop 1
Thought: I find the tomato, now I need to take it.\\
Action: take tomato 1}

In this example, the agent produces the action `\texttt{take tomato 1}' in response to the observation `\texttt{On the countertop 2, you see a tomato 1}'.

%% file: sections/4_experiments.tex
\section{Experimental setup}


Our aim is to study reasoning capabilities of LLM-based agents. In line with previous work, we use Alfworld \cite{shridhar2021alfworld}, a household robotics environment; Webshop \cite{yao2023webshop}, an online shopping environment; and Texcraft \citep{prasad2024adaptasneededdecompositionplanning}, a text-based tool crafting game, as our evaluation environments. 
We mainly \textbf{evaluate StateAct as the base agent}. However, we also show that StateAct can be used as a \textbf{drop-in replacement for ReAct} and validate it on test-time scaling. RAG (AutoGuide) and fine-tuning (ActRe)-based extensions are left for future work, as these are expensive and complex to run due to training on additional data and use of additional tools.


\subsection{Alfworld}
\begin{figure}[t]
  \includegraphics[width=1\columnwidth]{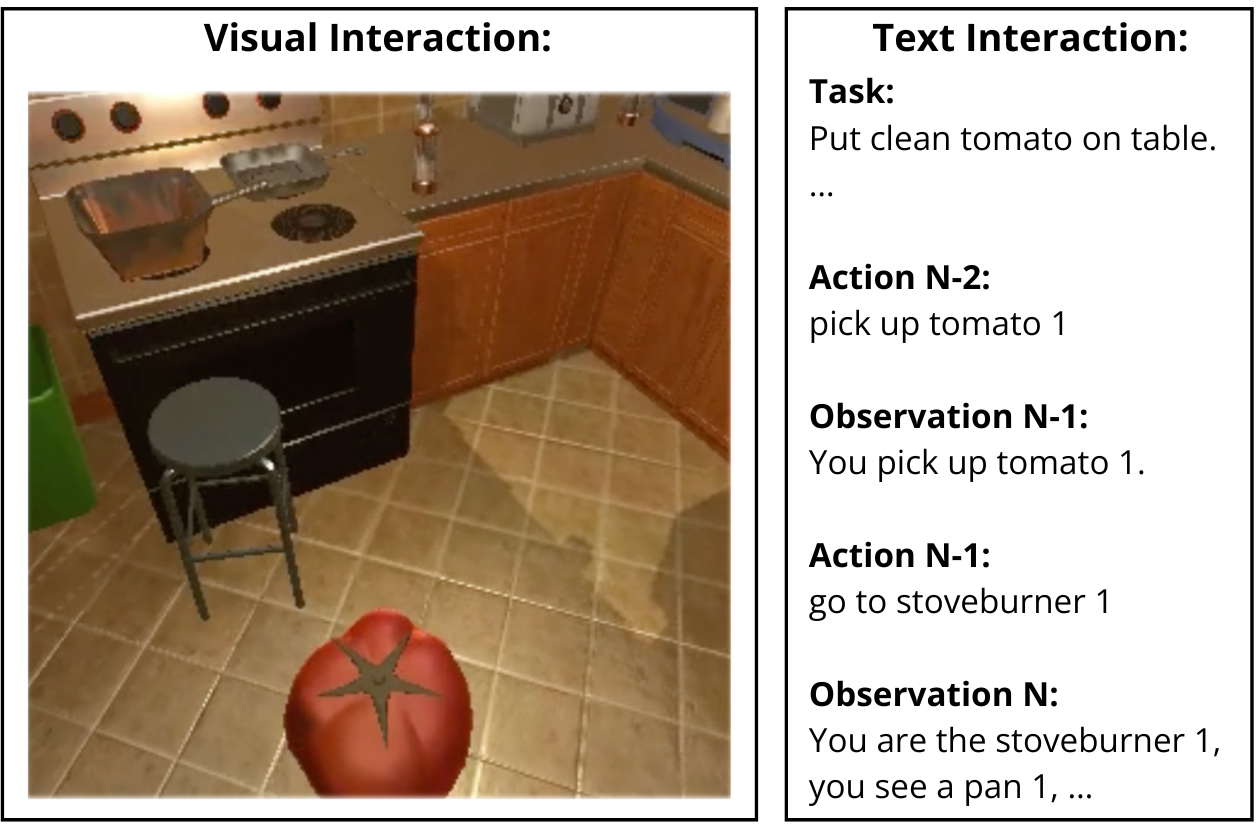}
  \caption{An example textual interaction in Alfworld (right) and corresponding 3D rendering (left).}
  \label{fig:alfworld_env}
\end{figure}
Alfworld \cite{shridhar2021alfworld} is based on a 3D visual household robotics environment called Alfred \cite{shridhar2020alfred}, which was translated into a text-based environment for ease of use for language-based AI models, see Figure \ref{fig:alfworld_env}. Alfworld has a total of 134 test-set examples and six environment types. It features long-time horizons, partial observability, an out-of-distribution evaluation set and text-based interactions. Alfworld simulates a household environment with a household assistant robot tasked with solving problems, e.g. \texttt{clean an apple and put it on a table}. The robot (or agent) then needs to perform a series of high-level operations to accomplish the tasks, e.g. `\texttt{go to fridge 1}', `\texttt{open fridge 1}'. At every step, the environment provides either a textual observation or the feedback that the command has failed, e.g. `\texttt{You open the fridge 1}', `\texttt{You see apple 1}'. The underlying text engine is based on Textworld \cite{côté2019textworld}, see Appendix \ref{sec:appendix_alfworld} for a complete list of the commands and details of the environments. 



\subsection{Webshop}
\begin{figure}[t]
  \includegraphics[width=1\columnwidth]{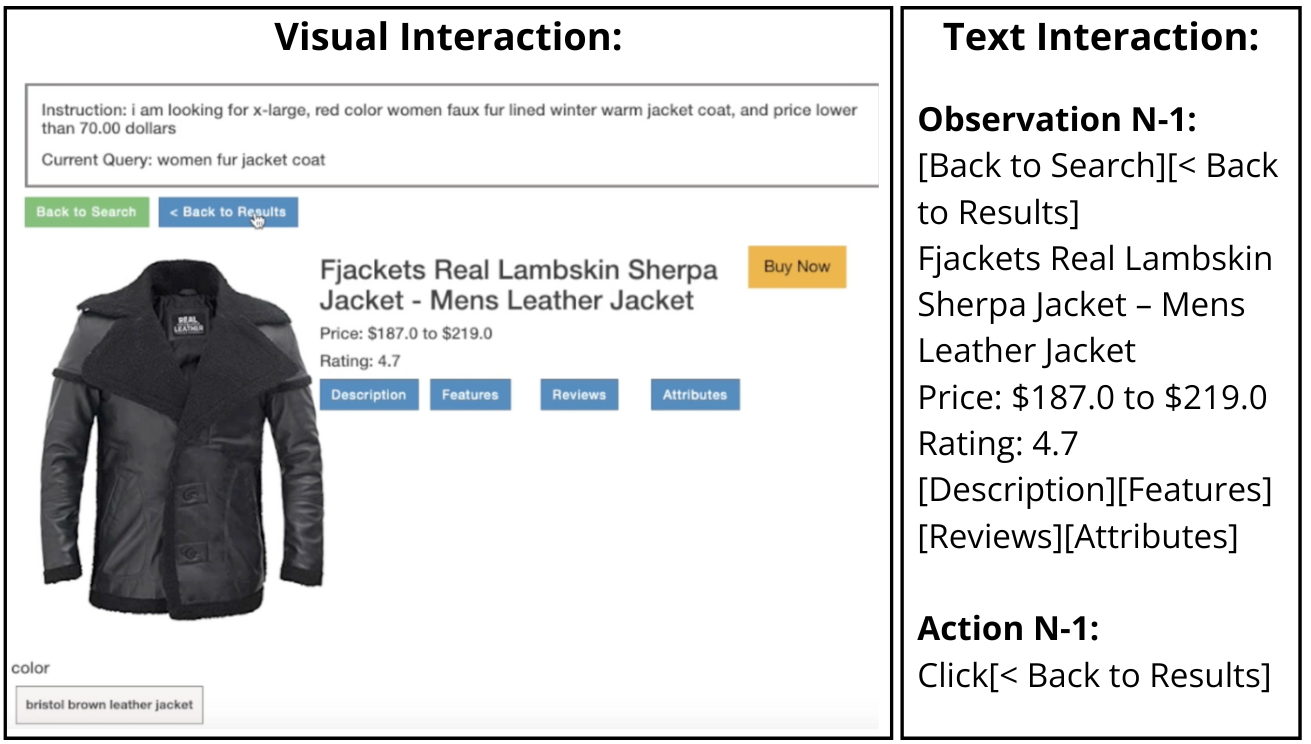}
  \caption{An textual interaction in Webshop (right) and correspondingwebsite rendering (left).}
  \label{fig:webshop_env}
\end{figure}
Webshop \cite{yao2023webshop} is a simulation of an online shopping experience. Given a task, e.g. ``\texttt{I want a blue water-proof winter jacket, less than \$100}'', the agent needs to search a product catalogue, browse through the search results, select the most fitting product, select the attributes, e.g. \texttt{colour}, \texttt{size}, and then buy the product. In line with previous work, we use the text-based version of Webshop, where all descriptions of the website are given in text form, see Figure \ref{fig:webshop_env}. Webshop features a realistic  large-scale product catalogue, a search engine, and very varied product attributes depending on the category of product, see Appendix \ref{sec:appendix_webshop} for more details. In total, the test set consists of 100 examples and each one is of the type ``search and buy a product''. Overall, Webshop has a maximum of fifteen steps and two commands: 1. \texttt{search[<query>]}, 2. \texttt{click[<button>]}.

\subsection{Textcraft}
Textcraft \cite{prasad2024adaptasneededdecompositionplanning}
is an environment based on the popular game Minecraft, where the task of the agent is to craft items. The environment is fully text-based. We use prompts and implementations based on the ADaPT paper \citep{prasad2024adaptasneededdecompositionplanning} that introduced this environment. We also use this environment to analyse whether StateAct performs well in combination with other methods such as ADaPT. See Appendix \ref{sec:appendix_textcraft} for more details.

\subsection{In-context learning}
Since ReAct \cite{yao2023react} forms the underlying agent for many current \cite{zhao2023expel} and state-of-the-art approaches \cite{Fu2024AutoGuideAG, yang2024react, prasad2024adaptasneededdecompositionplanning}, we use the same few-shot interaction traces as ReAct. The main reason is to have a fair comparison and to isolate additional effects, such as performance change, from different in-context examples. 
Alfworld, for example, has six types of tasks and ReAct uses two in-context examples per task type to prompt the language models. On average, each ReAct example ranges from 352 words to 591 words (590 tokens to 935 tokens). For our study, we reuse the observations, thoughts and actions, and annotate these examples further with the goal and state, which results in a range from 484 to 911 words (807 tokens to 1458 tokens) per example. During our annotation, we discovered minor errors in the ReAct prompts and fixed them as well. In comparison, AdaPlanner \cite{Sun2023AdaPlannerAP}, uses a different code-based approach and the prompt has 1104 words (2015 tokens) on average. 
We use the two-shot examples from ReAct for Alfworld, the one-shot example from ReAct for Webshop and the few-shot prompts of ReAct from ADaPT for Textcraft in all our experiments.

\subsection{Models}

In our work, we compare our method using newer state-of-the-art models, architectures and sizes to show that our method generalises. Specifically, we use Mistral-Small-24B-Instruct \citep{jiang2023mistral7b}, Qwen-2.5-7B,14B,32B-Instruct \citep{qwen2025qwen25technicalreport} and Gemma2-27B-Instruct \citep{gemmateam2024gemmaopenmodelsbased}.
We note that api-based models such as OpenAI's models are generally very expensive\footnote{A single evaluation run on Alfworld costs approx. \$8 using gpt-3.5 and ReAct, whilst gpt-4 costs 10+ times more.} and lack rigorous reproducibility standards\footnote{Since we do not have access to weights and inference settings and models become regularly deprecated.}. Nonetheless, we include experiments using \texttt{gpt-3.5} and \texttt{gpt-4o-mini}.
We use temperature 0 for all experiments and sample only the top 1 response; we use vllm for inference \citep{kwon2023efficientvllm}, see Appendix \ref{sec:appendix_compute} \& \ref{sec:appendix_code} for details.

%% file: sections/5_results.tex
\section{Results}
Overall we present results of state-of-the-art methods such as ActRe and ADaPT, as well as the underlying base agent ReAct. In Table \ref{tab:results_react} we see that while methods that rely on ReAct + extensions, such as ActRe and AutoGuide outperform StateAct overall, they also rely on additional training data and computation. Furthermore, StateAct achieves the best result among base agents, outperforming ReAct between 7\% and 30\% and achieves comparable result with the state-of-the-art methods, while not using any additional tools or data. For example, AutoGuide (ReAct + RAG) gets 0.79, while StateAct gets 0.77.
\begin{table}[h!]
    \centering
    \begin{tabular}{|l|c|}
        \hline
        \textbf{Method} & \textbf{Score} \\
        \hline
        State-of-the-art methods &  \\
        \hline
        AdaPlanner (Code-prompt + exec.)$^1$ & 0.75 \\
        AutoGuide (ReAct + RAG)$^2$ & 0.79 \\
        ActRe (ReAct + fine-tuning)$^3$ & \textbf{0.83} \\
        ADaPT (ReAct + test time scaling)$^4$ & 0.72 \\
        \hline
        Base agents &  \\
        \hline
        Act (few-shot only) & 0.41\\
        ReAct (few-shot only) & 0.64 \\
        AdaPlanner (Code-prompt only)$^1$ & 0.45 \\
        \textbf{StateAct} (\textbf{ours}, few-shot only) & \textbf{0.77} \\
        \hline
    \end{tabular}
    \caption{Results on the 134 test samples of Alfworld using gpt-3.5. ReAct and StateAct scores are single run with greedy decoding and gpt-3.5-1106. $^1$=code-execution \citep{Sun2023AdaPlannerAP}, $^2$=\citep{Fu2024AutoGuideAG}, $^3$=\citep{yang2024react}, $^4$=\citep{prasad2024adaptasneededdecompositionplanning}.}
    \label{tab:results_react}
\end{table}

\subsection{Base agent comparison}
\begin{table*}[h!]
    \centering
    \begin{tabular}{|l|ccccc|c|}
        \hline
        \textbf{Agent Name} & \textbf{Mistral-24B} & \textbf{Qwen-7B} & \textbf{Qwen-14B} & \textbf{Qwen-32B} & \textbf{Gemma-27B} & \textbf{Average} $\uparrow$\\
        \hline
        \textbf{Alfworld*}&\multicolumn{5}{l|}{}& \\
        \hline
        ReAct   & 0.44 & 0.10 & 0.75 & 0.89 & \textbf{0.71} & 0.58 \\
        StateAct & \textbf{0.49} & \textbf{0.46} & \textbf{0.78} & \textbf{0.90} & 0.68 & \textbf{0.66} \\
        \hline
        \textbf{Webshop**}&\multicolumn{5}{l|}{}& \\        
        \hline
        ReAct   & 0.34 & \textbf{0.19} & 0.22 & 0.27 & 0.26 & 0.26 \\
        StateAct & \textbf{0.35} & 0.12 & \textbf{0.33} & \textbf{0.32} & \textbf{0.29} & \textbf{0.28} \\
        \hline
        \textbf{Textcraft***}&\multicolumn{5}{l|}{}& \\        
        \hline
        ReAct   & 0.33 & 0.02 & 0.31 & 0.31 & 0.18 & 0.23 \\
        StateAct & \textbf{0.40} & \textbf{0.04} & \textbf{0.37} & \textbf{0.40} & \textbf{0.34} & \textbf{0.31} \\
        \hline
    \end{tabular}
    \caption{Base agent performance across different models and environments.
    *=134 Test Environments from Alfworld. 
    **=100 Test Environments from Webshop. 
    ***=100 Test Environments from Textcraft. 
    M=Mistral-Instruct-2501, Q=Qwen2.5-Instruct, G=Gemma 2-Instruct. 
    We use greedy decoding (temperature=0).}
    \label{tab:main_results}
\end{table*}

Since ReAct is the previous best base agent and forms the basis of state-of-the-art approaches, we compare against it in detail. The results in Table \ref{tab:main_results} show that StateAct outperforms ReAct across three different benchmarks and five different models. Sometimes the difference is quite substantial, with StateAct outperforming ReAct by more than 10 points. Across all 15 experiments only in two ReAct performs better than StateAct: in Webshop, where the 7B model is likely overwhelmed by the amount of textual input it receives, as Webshop has a lot of verbal input; and in the case of Gemma2-27B ReAct slightly outperforms StateAct on Alfworld; a hypothesis is that Gemma has a limited context length of 8192, while Alfworld requires long traces due to the longer step length.

We further validate our results with additional LLMs that are too large to fit on a single GPU or are closed source. In Table \ref{tab:results_additional} we see a significant performance increase. Using gpt-3.5, gpt-4o-mini and Mixtral-8x22B\footnote{Mixtral-8x22b-instruct-v0.1 was queried using Nvidia's NIM API \url{https://developer.nvidia.com/nim} [Last Accessed March 2025].} \citep{jiang2024mixtral} on Alfworld, ReAct achieves 63.7, 68.15 and 72.59, while StateAct achieves 77.04 (+13.3), 71.85(+3.7) and 83.70(+11.2) respectively.

\begin{table}[h]
\begin{tabular}{|l | l | c |} 
 \hline
\textbf{Method}&\textbf{Model}&\textbf{Success Rate \%}\\
\hline
ReAct &Gpt-3.5&0.64\\
StateAct &Gpt-3.5&\textbf{0.77} \\
\hline
ReAct &Gpt-4o-mini&0.68\\
StateAct &Gpt-4o-mini&\textbf{0.72}  \\
\hline
ReAct &Mixtral-8x22B&0.73\\
StateAct &Mixtral-8x22B&\textbf{0.84}\\
\hline
\end{tabular}
\caption{Results on the 134 test examples from Alfworld. Results are single run and greedy. Models used: gpt-3.5-1106, gpt-4o-2024-07-18, mixtral-8x22b-instruct-v0.1. 
}
\label{tab:results_additional}
\end{table}

\subsection{Base agents + test time scaling}
An important contribution is to validate that StateAct can be used as a drop-in replacement with advanced methods. To this end we validate StateAct using test-time scaling using the ADaPT method. Starting with ADaPT, which is based on ReAct, as the starting point\citep{prasad2024adaptasneededdecompositionplanning}, we enhance their method using StateAct. In Table \ref{tab:results_test_time_scaling_adapt} we can clearly see that StateAct scales well with test time scaling jumping in performance by 39\%, and surpassing ADaPT+ReAct by 12\%.

\begin{table}[!h]
\centering
\begin{tabular}{|l|c|c|}
\hline
\textbf{Model/Agent Name} & \textbf{Normal} & \textbf{+ADaPT} \\
\hline
\textit{Mistral-24B} &  &  \\
\hline
ReAct & 0.33 & 0.53 \\
StateAct & \textbf{0.40} & \textbf{0.64} \\
\hline
\textit{Qwen2.5-7B} & & \\
\hline
ReAct  & 0.02 & 0.09 \\
StateAct & \textbf{0.04} & \textbf{0.11} \\
\hline
\textit{Qwen2.5-14B} & & \\
\hline
ReAct  & 0.31 & \textbf{0.55} \\
StateAct & \textbf{0.37} & 0.53 \\
\hline
\textit{Qwen2.5-32B} & & \\
\hline
ReAct & 0.31 & \textbf{0.64} \\
StateAct & \textbf{0.40} & 0.62 \\
\hline
\textit{Gemma-2-27B} & & \\
\hline
ReAct  & 0.18 & 0.19 \\
StateAct & \textbf{0.34} & \textbf{0.35} \\
\hline
\textit{\textbf{Average}} & & \\
\hline
ReAct  & 0.23 & 0.4 \\
StateAct & \textbf{0.31} & \textbf{0.45} \\
\hline
\end{tabular}
\caption{Comparison of test-time scaling performance on 100 test samples from Textcraft. Normal refers to using just the base agent. +ADaPT means running the respective base agent with test time scaling using the ADaPT method. ADaPT code is adapted to run ReAct and StateAct. We use greedy decoding and $d_{max}$ = 2.}
\label{tab:results_test_time_scaling_adapt}
\end{table}


\subsection{Summary of results}
StateAct establishes itself as the best-performing base agent, surpassing ReAct by 7–30\% across multiple benchmarks while requiring no additional training data or external tools. Although advanced methods like ActRe and AutoGuide achieve higher scores, they rely on costly training and retrieval.  We also validate StateAct as a drop-in replacement for ReAct in test-time scaling. These results highlight the performance and usability of StateAct as a robust foundation for LLM-based agents.

%% file: sections/6_analysis.tex
\section{Analysis and Ablations}
In the results section, we discovered that our methods outperform the previous state-of-the-art base agent, ReAct. Our method shows strong performance with in-context learning without resorting to additional tools or data. In this section, we analyse our results further and also show that \textit{self prompting} and \textit{state-tracking} help with long-range reasoning. For most ablation studies, we focus on Alfworld as it has two favourable properties over Webshop and Textcraft. Firstly, Alfworld has a longer time horizon (50 steps vs. 15 in Webshop, 40 in Textcraft). Secondly, Alfworld is more realistic than Textcraft, as Alfworld is a robotics environment, while Textcraft is based on a game. For completeness, we include ablation results in Table \ref{tab:abltation_table} and full results in Appendix \ref{appendix:additional_alfworld},\ref{appendix:additional_webshop},\ref{appendix:additional_textcraft}.
\subsection{Does self-prompting help with long-range tasks?}
For this purpose, we compare the original ReAct (thought + action) with the self-prompting included, i.e. StateAct (goal + thought + action). In Figure \ref{fig:abl_state} we can see that, while the performance of both ReAct and StateAct goes down as there are more steps, the goal tracking has better relative performance as the number of steps increases and is able to solve longer tasks of 40 to 50 steps. This finding is in line with our original motivation, that LLM agents deteriorate in performance as the prompts and interactions get longer.

To verify that this actually means that goal tracking helps with performance, as opposed to just increasing the number of steps it takes to solve a task, we calculate the average number of steps for ReAct\footnote{We ignore `thought' turns for ReAct as otherwise ReAct would have even more steps.} and StateAct. Table \ref{tab:ablation_steps} clearly shows that ReAct, with an average of 31.49 steps to solve an environment, is the least efficient whilst StateAct, with an average of 19.11 steps to solve an environment, is the most efficient. This shows that not only does self-prompting help with longer range tasks, it also help with efficiency, by shortening the tasks. See Appendix \ref{sec:appendix_analysis} for more discussion.
\subsection{What effects does state-tracking have?}
\begin{figure}[t]
\centerline{\includegraphics[width=1\columnwidth]{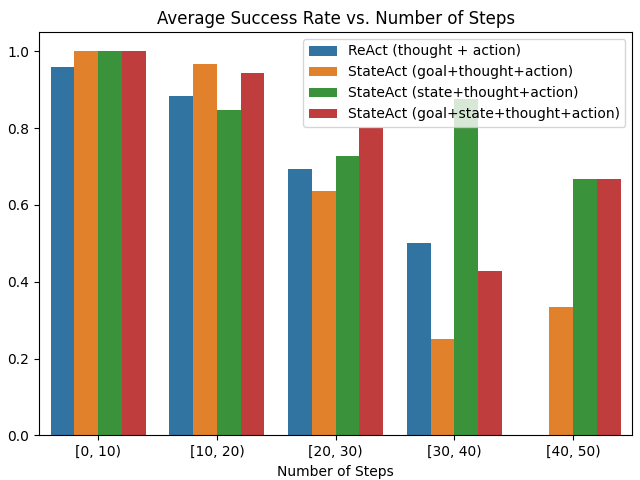}}
\caption{State vs. No State, on the 134 test examples from Alfworld, using gpt-3.5-turbo-1106
}
\label{fig:abl_state}
\end{figure}

We also analyse whether state-tracking helps with long-range reasoning and efficiency. We compare the full StateAct against StateAct without state-tracking, as well as comparing ReAct (thought + action) against StateAct with state-tracking (state + thought + action).
In Figure \ref{fig:abl_state} we see that state-tracking also helps with long-range reasoning. In fact, we can see that reasoning alone is unable to solve tasks longer than 40 steps, while with both state tracking
and goal-tracking longer-range tasks can be solved.  Also, looking at Table \ref{tab:ablation_steps} we see that state-tracking makes the model the most efficient\footnote{Despite StateAct using a twice-longer prompt, our cost remains similar to ReAct, at around \$8 for the full Alfworld run, since we solve tasks more efficiently and use fewer steps.}. Concretely, ReAct has 24/134 examples that are in the bucket `40-50’ and solves 0, while StateAct has 6/134 examples and solves 4. Therefore, we find that explicit state-tracking helps with long-range tasks and to solve the tasks more efficiently.

\begin{table}[t]
\begin{tabular}{|l | c |} 
 \hline
\textbf{Method}&\textbf{Avg. Steps $\downarrow$}\\
\hline
ReAct &31.49\\
StateAct (goal, state, thought)&\textbf{19.11}\\
- w/o goal (self-prompt) &20.09\\
- w/o state & 22.50\\
- w/o thought&23.76\\
\hline
\end{tabular}
\caption{Average number of steps (Avg. Steps) [lower is better] on the test set of Alfworld, using gpt-3.5-1106.}
\label{tab:ablation_steps}
\end{table}

\subsection{Does the model perform actual state-tracking?}
\begin{figure}[t]
\centerline{\includegraphics[width=1\columnwidth]{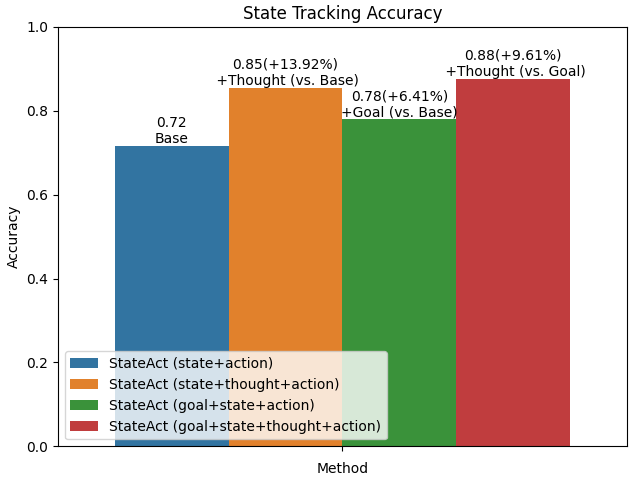}}
\caption{State-tracking accuracy for StateAct on 134 test examples of Alfworld, using gpt-3.5-1106.}
\label{fig:ablation_correct_states}
\end{figure}

We investigate whether the model is actually performing state-tracking. For that purpose, we look at Alfworld and construct a self-verification algorithm that is able to track the state heuristically\footnote{See Appendix \ref{sec:appendix_heuristic} for details.} based on the actions the agent takes. For example, if the agent produces the action \texttt{go to fridge 1} and the environment accepts this action, we update the state with \texttt{current location: fridge 1}. We compare the \textit{`gold' state} against the predicted state. Figure \ref{fig:ablation_correct_states} shows that StateAct achieves a state-tracking accuracy of 88\%. We also observe that thoughts and goals help the state-tracking.

\subsection{Do `thoughts' help?}
While chain-of-thought helps in most cases, surprisingly, we see that in some cases chain-of-thought can be harmful. Specifically, in Webshop `thoughts' actually harm overall performance across different agent and model types, see Table \ref{tab:abltation_table}. A hypothesis is the verbosity of the Webshop environment, thus confusing the model with `verbal' thoughts.

\begin{table}[h]
\centering
\begin{tabular}{|l|ccc|}
\hline
\textbf{Agent Name} & \textbf{AW} & \textbf{WS} & \textbf{TC} \\
\hline
\textbf{Baselines}&\multicolumn{3}{l|}{} \\
\hline
Act & 0.51 & 0.19 & 0.27 \\
Thought+Act (ReAct)  & 0.58 & 0.26 & 0.23 \\
\hline
\textbf{Our Methods}&\multicolumn{3}{l|}{} \\
\hline
State+Act & 0.51 & \textbf{0.31} & 0.22 \\
State+Thought + Act & 0.58 & 0.18 & \textbf{0.34} \\
\hline
Goal+Act & 0.63 & 0.24 & 0.26 \\
Goal+Thought + Act & 0.65 & 0.18 & 0.30 \\
\hline
Goal+State+Act & 0.56 & 0.29 & 0.21 \\
Goal+State+Thought+Act& \textbf{0.66} & 0.28 & 0.31 \\
\hline
\end{tabular}
\caption{Ablation table. Averaged results across Mistral, Qwen, Gemma on AW=Alfworld, WS=Webshop and TC=Textcraft. Goal=Self-prompt.
}
\label{tab:abltation_table}
\end{table}

%% file: sections/7_conclusion.tex
\section{Conclusion}
Our work is driven by the fundamental challenge that LLM agents struggle with long context and keeping on track with instructions. The current state-of-the-art to overcome such challenges propose extensions on top of the base agent, ReAct. In contrast, we introduced a novel base agent, StateAct, an in-context learning method that leverages \textit{chain-of-states} and \textit{self-prompting} to significantly enhance the capabilities of LLM agents. By rethinking how agents track and utilize state information, StateAct establishes a new state-of-the-art for base agents, surpassing ReAct by 9\% to 20\% across different models and tasks. Furthermore, we demonstrate that StateAct scales effectively when combined with advanced techniques such as ADaPT, reinforcing its usability.

Beyond raw performance, our analysis uncovers a crucial insight: StateAct not only improves reasoning but also enhances efficiency, allowing agents to achieve better results with fewer steps. This suggests that integrating structured state-tracking and self-prompting cues helps mitigate the well-documented long-context issue in LLMs.

Our findings suggest that methods like StateAct can serve as a practical and efficient way to improve LLM-based agents without requiring costly external augmentations or data such as retrieval or fine-tuning. By enabling agents to manage their own state explicitly, StateAct provides a scalable approach to improving reasoning and decision-making across a range of tasks. This makes StateAct a viable drop-in solution for current systems, offering both better performance and greater efficiency in LLM agent tasks.

%% file: sections/8_ethics.tex
\section{Ethical Considerations}
\subsection{Computational footprint}
Running many of the experiments presented in this paper can have a significant computational footprint. We should consider the environment and financial resources for reproduciblity of our work. We aimed to address this concern by models that are less computationally demanding such as \texttt{gpt-3.5-turbo} level models or open source models that fit on a single GPU (A100, 80GB), reporting costs and minimising the cost of our method.

\subsection{Hallucinations in LLMs}
As LLM-based agents become more powerful and therefore more pervasive in our daily lives, `hallucinations' in LLMs can be very harmful \cite{wei2024measuring}. We hope that explicit state-tracking presented in this work can also lead to future work to reduce `hallucinations.'

%% file: sections/9_limitations.tex
\section{Limitations}
\subsection{Languages and evaluation benchmarks}
We evaluated our method only in the English language and on three evaluation benchmarks. While we do not expect major changes in other languages, this is something that should be investigated.

\subsection{Reasoning traces that rely on human judgement}
Our prompts require human annotations; as such, there is a natural bias present. This can have both task-performance implications as well as ethical implications.


%% file: sections/appendix.tex
\newpage
\appendix

\section{Alfworld}
\label{sec:appendix_alfworld}
\subsection{Environment Types}
Alfworld has six different environment types:
1. \textit{clean}, 2. \textit{heat}, 3. \textit{cool}, 4. \textit{examine}, 5. \textit{put}, 6. \textit{puttwo}.

The `\textit{clean}' task, e.g. \texttt{Task: Put a clean apple on table}, requires the agent to first find the apple, then clean it (in the sink/basin) and then put it on a table.

The `\textit{heat}' task, e.g. \texttt{Task: Put a hot pie on table}, requires the agent to first find the pie, then heat it (on the stove/burner) and then put it on a table.

The `\textit{cool}' task, e.g. \texttt{Task: Put a cool tomato on table}, requires the agent to first find the tomato, then cool it (with the fridge) and then put it on a table.

The `\textit{examine}' task, e.g. \texttt{Task: Examine the mug with the desklamp}, requires the agent to first find the mug, then find the desk lamp, and then use the desk lamp.

The `\textit{put}' task, e.g. \texttt{Task: Find some apple and put it in sidetable}, requires the agent to first find an apple, and then put it on the side table.

The `\textit{puttwo}' task, e.g. \texttt{Task: Put two cellphone in sofa}, requires the agent to first find one cellphone, and then put it on the sofa, and then to find the second one and put it on the sofa.
\subsection{Action Types}
Alfworld has the following valid actions:
1. \textit{go to}, 2. \textit{open}, 3. \textit{close}, 4. \textit{put}, 5. \textit{take}, 6. \textit{cool}, 7. \textit{heat}, 8. \textit{use}.

\texttt{
go to <place> \\ Example: go to table 1\\\\
open <object> \\ Example: open door 1\\\\
close <object> \\ Example: close door 1\\\\
put <object> in/on <place> \\ Example: put apple 1 in/on table 1\\\\
take <object> from <place> \\ Example: take apple 1 from table 1\\\\
cool <object> with <place> \\ Example: cool apple 1 with fridge 1\\\\
heat <object> with <place> \\ Example: heat apple 1 with fire 1\\\\
use <object> \\ Example: use desklamp 1}

\subsubsection{Alfworld correction} 
\label{sec:alfworld_correction}
In our research, we identified that Alfworld has a specific syntactic feature for the \texttt{put} command, namely \texttt{put <object> in/on <place>}, where ``in/on'' needs to be written exactly this way. Using only ``in'' or only ``on'' produces a failed command. 
We observed this issue with LLMs in this environment and we propose a simple fix for it. 
We map: 1. ``\texttt{put <object> in <place>}'' and 2. ``\texttt{put <object> on <place>}'' to the command accepted by Alfworld, namely ``\texttt{put <object> in/on <place>}''.

\textbf{Note:} In the latest release of Alfworld (December 2024) this was fixed by replaceing the \texttt{put <obj> in/on <place>} command with \texttt{move <obj> to <place>} command. In our work we report results on the latest version of alfworld.

\subsection{License}
Alfworld has the permissible MIT license; we used it in line with the license.

\section{Webshop}
\label{sec:appendix_webshop}

\subsection{Commands and environment}
Webshop has one environment type: `\textit{search \& buy}', as well as two commands: 1. \textit{search}, 2. \textit{click}.

\texttt{click[<button>] \\ Example: click[< Back to Search]\\\\
search[<query>] \\ Example: search[interesting book]}

\subsection{Prodcuts and attributes}
Webshop has over 1 million real-world products across 5 main categories (fashion, makeup, electronics, furniture and food) and 113 sub-categories.

\subsection{License}
Webshop has the permissible Princeton license; we used it in line with the license.

\section{Textcraft}
\label{sec:appendix_textcraft}

\subsection{Commands and environment}
Textcraft has one environment type: `\textit{craft}', as well as three commands: 1. \textit{inventory}, 2. \textit{craft}, 3. \textit{get}

\subsection{Crafting Recipes}
Textcraft has crafting recipes that range from easy to hard. Where hardness is measured by the `depth' of the crating recipe. Specifically, depths of 2, 3 and 4 are present in the dataset.

\subsection{License}
Textcraft is published under the permissible MIT license. 

\section{Compute Requirements for local LLMs}
\label{sec:appendix_compute}
The exact code will be released upon publication. However, to help reproducibility we ran all experiments on single A100 80GB GPUs. In terms of software we used: vLLM for inference. The hyper-parameters were set to: max model length 16000 (except for Gemini, where we used 8192), temperature = 0, datatype="auto" (which results in bfloat16).

\section{Code} 
\label{sec:appendix_code}
\subsection{Code Snippet to call Local LLMs}
\label{sec:appendix_code_local}
\begin{lstlisting}
from vllm import LLM, SamplingParams
self.llm = LLM(
    model=model,
    tensor_parallel_size=tensor_parallel_size,
    gpu_memory_utilization=0.95,
    max_model_len=max_model_len,
    dtype="auto"
)
        
messages = [
    # {"role": "system", "content": "You are a helpful assistant."},
    {"role": "user", "content": prompt[-self.max_model_len:]}
]
text = self.tokenizer.apply_chat_template(
    messages,
    tokenize=False,
    add_generation_prompt=True
)

sampling_params = SamplingParams(
    temperature=self.temperature,
    top_p=1.0,
    repetition_penalty=1.00,
    max_tokens=min(2000,self.max_model_len),
    stop = self.stop_sequences,
    seed = self.seed
)
outputs = self.llm.generate([text], sampling_params)
return outputs[0].outputs[0].text
\end{lstlisting}
\subsection{Code Snippet to call OpenAI / GPT-3.5}
\label{sec:appendix_code_api}
\begin{lstlisting}
client = openai.OpenAI(
        # Defaults to os.environ.get("OPENAI_API_KEY")
        # api_key=OPENAI_KEY,
)
        
full_prompt = [{
    "role": "user", 
    "content": prompt
}]

chat_completion = client.chat.completions.create(
    model="gpt-3.5-turbo-1106",
    messages=full_prompt,
    temperature=0.0,
    stop = ["\n\n"]
)
\end{lstlisting}

A prompt is given in Appendix \ref{sec:appendix_prompts}.

\section{Heuristic State-tracking Explained}
\label{sec:appendix_heuristic}
Heuristic state tracking is based on the idea that the state can be inferred automatically if one follows the actions of the agent and observations of the environments. Specifically, the state at time $t$ for StateAct depends on the state at time \texttt{t-1} and the action $a_{t-1}$ and observation $o_t$. For example, if the state at time $t$ the `current location' of the state is set to \texttt{table 1} and the action is \texttt{go to fridge 1} and the observation is successful, then the `current location' can be updated to be \texttt{fridge 1} automatically. This rule based `state-tracking' is how the heuristics work. 

\section{`Step Length Analysis' Discussion}
\label{sec:appendix_analysis}
An alternative to calculating and comparing step length could be `gold solutions' to measure optimal step length and optimality of an agent. We see two issues. Firstly, the annotation cost of creating gold solutions. Secondly, it is not clear what the gold solution should be. Concretely in Alfworld, an `oracle’ solution could have very few steps as it would immediately go to the location of the `hidden’ object, while a `non-oracle’ expert solution would have more steps as more locations would be searched. Thus a `difficulty’ measure would be needed instead, but it is ambiguous.

\section{Potential Future Work Directions}
We found that `thoughts' or explicit reasoning do not always help performance. It would be very interesting to systematise `thought' and `states' and to understand what contributes positively and the reasons why. Also, inspired by the positive results of StateAct, it is interesting to see what other improvements can be made without resorting to training, larger models or external tools. Finally, problems related to \textit{domain-specific} syntax are also an interesting avenue for future work.

\section{Does JSON structure help StateAct performance?}

We also investigated whether adding a structured format like json would help. For this purpose, we re-ran StateAct on Alfworld, but translated the state into a json format, see section \ref{sec:appendix_json} for more details. Surprisingly, we found that the json format hinders performance, see Table \ref{tab:ablation_json}. 

\begin{table}[h]
\begin{tabular}{|l | c | c |} 
 \hline
\textbf{Method}&\textbf{SR\%} &\textbf{SR (+json)\%}\\
\hline 
ReAct &\textbf{63.70}& 62.96(-0.74)\\
StateAct (complete) &\textbf{77.04}&58.52(-18.5)\\
\hline
\end{tabular}
\caption{No-json vs. json. Success Rate (SR) on the test set of Alfworld, using gpt-3.5-1106.}
\label{tab:ablation_json}
\end{table}

\subsection{JSON prompt}
\label{sec:appendix_json}
We translate the text-based StateAct prompt:
\begin{lstlisting}
>goal: put a hot apple in fridge
current location: starting location
current inventory: None
thought: To solve the task, I need to find and take an apple, then heat it with microwave, then put it in fridge. First I need to find an apple. An apple is more likely to appear in fridge (1), diningtable (1), coffeetable (1), drawer (1), cabinet (1-13), garbagecan (1). I can check one by one, starting with fridge 1.
action: go to fridge 1
\end{lstlisting}

Into the following json format:
\begin{lstlisting}
>{"goal": "put a hot apple in fridge",
"current_location": "starting location",
"current_inventory": None,
"thought": "To solve the task, I need to find and take an apple, then heat it with microwave, then put it in fridge. First I need to find an apple. An apple is more likely to appear in fridge (1), diningtable (1), coffeetable (1), drawer (1), cabinet (1-13), garbagecan (1). I can check one by one, starting with fridge 1."
"action": "go to fridge 1"
}
\end{lstlisting}

\section{StateAct Prompts}
\label{sec:appendix_prompts}
\subsection{Example Alfworld prompt for the `heat' task}
\begin{lstlisting}
Interact with a household to solve a task.

Here are 2 examples:

You are in the middle of a room. Looking quickly around you, you see a cabinet 13, a cabinet 12, a cabinet 11, a cabinet 10, a cabinet 9, a cabinet 8, a cabinet 7, a cabinet 6, a cabinet 5, a cabinet 4, a cabinet 3, a cabinet 2, a cabinet 1, a coffeemachine 1, a countertop 1, a diningtable 1, a drawer 1, a fridge 1, a garbagecan 1, a microwave 1, a shelf 3, a shelf 2, a shelf 1, a sinkbasin 1, a stoveburner 4, a stoveburner 3, a stoveburner 2, a stoveburner 1, and a toaster 1.
Your task is to: put a hot apple in fridge.

>goal: put a hot apple in fridge
current location: starting location
current inventory: None
thought: To solve the task, I need to find and take an apple, then heat it with microwave, then put it in fridge. First I need to find an apple. An apple is more likely to appear in fridge (1), diningtable (1), coffeetable (1), drawer (1), cabinet (1-13), garbagecan (1). I can check one by one, starting with fridge 1.
action: go to fridge 1

The fridge 1 is closed.

>goal: put a hot apple in fridge
current location: fridge 1
current inventory: None
thought: None
action: open fridge 1

You open the fridge 1. The fridge 1 is open. In it, you see a cup 1, and a egg 1.

>goal: put a hot apple in fridge
current location: fridge 1
current inventory: None
thought: None
action: go to diningtable 1

On the diningtable 1, you see a apple 1, a bread 1, a fork 3, a fork 2, a fork 1, a mug 2, a peppershaker 3, a plate 2, a pot 1, a soapbottle 3, a spatula 1, a spoon 3, a spoon 2, a spoon 1, a winebottle 3, and a winebottle 2.

>goal: put a hot apple in fridge
current location: diningtable 1
current inventory: None
thought: Now I find an apple (1). Next, I need to take it.
action: take apple 1 from diningtable 1

You pick up the apple 1 from the diningtable 1.

>goal: put a hot apple in fridge
current location: diningtable 1
current inventory: apple 1
thought: Now I take an apple (1). Next, I need to go to a microwave (1) and heat it.
action: go to microwave 1

The microwave 1 is closed.

>goal: put a hot apple in fridge
current location: microwave 1
current inventory: apple 1
thought: None
action: heat apple 1 with microwave 1

You heat the apple 1 using the microwave 1.

>goal: put a hot apple in fridge
current location: microwave 1
current inventory: apple 1
thought: Now I heat an apple (1). Next, I need to put it in/on fridge 1.
action: go to fridge 1

The fridge 1 is open. In it, you see a cup 1, and a egg 1.

>goal: put a hot apple in fridge
current location: fridge 1
current inventory: apple 1
thought: None
action: put apple 1 in/on fridge 1

You are in the middle of a room. Looking quickly around you, you see a cabinet 10, a cabinet 9, a cabinet 8, a cabinet 7, a cabinet 6, a cabinet 5, a cabinet 4, a cabinet 3, a cabinet 2, a cabinet 1, a coffeemachine 1, a countertop 3, a countertop 2, a countertop 1, a diningtable 1, a drawer 6, a drawer 5, a drawer 4, a drawer 3, a drawer 2, a drawer 1, a fridge 1, a garbagecan 1, a microwave 1, a sinkbasin 1, a stoveburner 4, a stoveburner 3, a stoveburner 2, a stoveburner 1, and a toaster 1.
Your task is to: heat some egg and put it in diningtable.

>goal: heat some egg and put it in diningtable
current location: starting location
current inventory: None
thought: To solve the task, I need to find and take an egg, then heat it with microwave, then put it in diningtable. First I need to find an egg. An egg is more likely to appear in fridge (1), countertop (1-3), diningtable (1), stoveburner (1-4), toaster (1), garbagecan (1), cabinet (1-10). I can check one by one, starting with fridge 1.
action: go to fridge 1

The fridge 1 is closed.

>goal: heat some egg and put it in diningtable
current location: fridge 1
current inventory: None
thought: None
action: open fridge 1

You open the fridge 1. The fridge 1 is open. In it, you see a lettuce 2, a mug 2, and a potato 3.

>goal: heat some egg and put it in diningtable
current location: fridge 1
current inventory: None
thought: None
action: go to countertop 1

On the countertop 1, you see a bread 1, a fork 1, and a saltshaker 1.

>goal: heat some egg and put it in diningtable
current location: countertop 1
current inventory: None
thought: None
action: go to countertop 2

On the countertop 2, you see nothing.

>goal: heat some egg and put it in diningtable
current location: countertop 2
current inventory: None
thought: None
action: go to countertop 3

On the countertop 3, you see a bowl 1, a butterknife 1, a egg 2, a kettle 2, a plate 1, a sink 1, and a spatula 2.

>goal: heat some egg and put it in diningtable
current location: countertop 3
current inventory: None
thought: Now I find an egg (2). Next, I need to take it.
action: take egg 2 from countertop 3

You pick up the egg 2 from the countertop 3.

>goal: heat some egg and put it in diningtable
current location: countertop 3
current inventory: egg 2
thought: Now I take an egg (2). Next, I need go to a microwave (1) and heat it.
action: go to microwave 1

The microwave 1 is closed.

>goal: heat some egg and put it in diningtable
current location: microwave 1
current inventory: egg 2
thought: None
action: heat egg 2 with microwave 1

You heat the egg 2 using the microwave 1.

>goal: heat some egg and put it in diningtable
current location: microwave 1
current inventory: egg 2
thought: Now I heat an egg (2). Next, I need to put it in/on diningtable 1.
action: go to diningtable 1

On the diningtable 1, you see a apple 2, a bread 3, a egg 1, a kettle 1, a knife 1, a mug 1, a papertowelroll 1, a peppershaker 2, a potato 1, a soapbottle 1, and a spatula 1.

>goal: heat some egg and put it in diningtable
current location: diningtable 1
current inventory: egg 2
thought: None
action: put egg 2 in/on diningtable 1

Here is the task.
<CURRENT TASK>
\end{lstlisting}

\subsection{Example Webshop prompt}
\begin{lstlisting}
Webshop 
Instruction:  
i would like a 3 ounce bottle of bright citrus deodorant for sensitive skin, and price lower than 50.00 dollars 
[Search]  

Goal: Buy a 3 ounce bottle of bright citrus deodorant for sensitive skin, and price lower than 50.00 dollars
Current Location: Search Home Page
Current Selection: None
Thought: None
Action: search[3 ounce bright citrus deodorant sensitive skin]

Observation: 
[Back to Search] 
Page 1 (Total results: 50) 
[Next >] 
[B078GWRC1J] 
Bright Citrus Deodorant by Earth Mama | Natural and Safe for Sensitive Skin, Pregnancy and Breastfeeding, Contains Organic Calendula 3-Ounce 
$10.99 
[B078GTKVXY] 
Ginger Fresh Deodorant by Earth Mama | Natural and Safe for Sensitive Skin, Pregnancy and Breastfeeding, Contains Organic Calendula 3-Ounce 
$10.99 
[B08KBVJ4XN] 
Barrel and Oak - Aluminum-Free Deodorant, Deodorant for Men, Essential Oil-Based Scent, 24-Hour Odor Protection, Cedar & Patchouli Blend, Gentle on Sensitive Skin (Mountain Sage, 2.7 oz, 2-Pack) 
$15.95  

Goal: Buy a 3 ounce bottle of bright citrus deodorant for sensitive skin, and price lower than 50.00 dollars
Current Location: Search Results Page, page 1
Current Selection: None
Thought: B078GWRC1J and B078GTKVXY are bright citrus deodorant less then 50 dollars. I can check B078GWRC1J first.
Action: click[B078GWRC1J]

Observation: 
[Back to Search] 
[< Prev] 
scent [assorted scents][bright citrus][calming lavender][ginger fresh][simply non-scents]
size [travel set (4-pack)][3 ounce (pack of 1)][3-ounce (2-pack)]
Bright Citrus Deodorant by Earth Mama | Natural and Safe for Sensitive Skin, Pregnancy and Breastfeeding, Contains Organic Calendula 3-Ounce 
Price: $10.99 
Rating: N.A. 
[Description] 
[Features] 
[Reviews] 
[Buy Now]  

Goal: Buy a 3 ounce bottle of bright citrus deodorant for sensitive skin, and price lower than 50.00 dollars
Current Location: Item B078GWRC1J Page
Current Selection: None
Thought: For 3 ounce bottle of bright citrus deodorant for sensitive skin, the item has options 'bright citrus' and '3 ounce (pack of 1)' and seems good to buy.
Action: click[bright citrus]

Observation: You have clicked bright citrus. 

Goal: Buy a 3 ounce bottle of bright citrus deodorant for sensitive skin, and price lower than 50.00 dollars
Current Location: Item B078GWRC1J Page
Current Selection: bright citrus
Thought: None
Action: click[3 ounce (pack of 1)]

Observation: You have clicked 3 ounce (pack of 1). 

Goal: Buy a 3 ounce bottle of bright citrus deodorant for sensitive skin, and price lower than 50.00 dollars
Current Location: Item B078GWRC1J Page
Current Selection: bright citrus, 3 ounce (pack of 1)
Thought: None
Action: click[Buy Now]
\end{lstlisting}

\section{Formalizing StateAct}
\label{sec:appendix_formalizing}
Let us denote by $\pi$ the policy of an AI-agent, in the standard case at time step $t$, the policy predicts action $a_t$, given the history of observations and actions $[o_t,a_{t-1},...,a_0,o_0]$.
\begin{equation}
\pi(a_t | o_t,a_{t-1},...,a_0,o_0)
\label{eq:basic_agent}
\end{equation}
Where $a_t$ is the action produced by the agent at time step t and $o_t$ is the observation produced by the environment at time step t after receiving action $a_t$ as input. Usually, the first observation $o_0$ also contains the `goal' description for the given environment. 

For our case, we need to enhance the policy to incorporate the `state'. Similar to previous work \cite{yao2023react}, we introduce the \textit{context} vector, $c_t$. The context vector contains the action, as well as the additional predictions of the agent
, i.e. $c_t=(g_0,s_t,r_t,a_t)$.
Where $g_0$ is the goal and it always remains the same (for a given environment) and uses the goal extracted from $o_0$; $s_t$ represents the predicted state at time step $t$; $r_t$ represents `chain-of-thought'-style `reasoning' at time step $t$; and $a_t$ represents the action at time step $t$, as before. The new policy $\pi$ then becomes:
\begin{equation}
\pi_{contextual}(c_t | o_t,c_{t-1},...,c_0,o_0)
\label{eq:contextual_agent}
\end{equation}

In our case, the LLM acts as $\pi_{contextual}$ and produces the context vector at every time step.

\section{Additional results for Alfworld}
\label{appendix:additional_alfworld}
For Alfworld we can see that the full StateAct performs best. Interstingly, adding state in does not always help model performance. A potential explanation for this can be that state-tracking itself is not the challenge in this dataset. Rather self-prompting is critical in Alfworld. Which indicates that long-range reasoning depends on self-prompting.

\begin{table*}[h]
\centering
\begin{tabular}{|l|ccccc|c|}
\hline
\textbf{Agent Name} & \textbf{M-24B} & \textbf{Q-7B} & \textbf{Q-14B} & \textbf{Q-32B} & \textbf{G-27B} & \textbf{Average}\\
\hline
\textbf{Baselines}&\multicolumn{5}{l|}{\textbf{}} &\\
\hline
Act-only (action) & 0.29 & 0.31 & 0.66 & 0.87 & 0.41 & 0.51 \\
ReAct (thought + action) & 0.44 & 0.10 & 0.75 & 0.89 & 0.71 & 0.58 \\
\hline
\textbf{Our Methods}&\multicolumn{5}{l|}{\textbf{}} &\\
\hline
StateAct (state+action) & 0.09 & 0.57 & 0.75 & 0.76 & 0.40 & 0.51 \\
StateAct (self-prompt+action) & 0.36 & 0.52 & 0.77 & 0.90 & 0.62 & 0.63 \\
StateAct (state+react) & 0.21 & 0.44 & 0.76 & 0.77 & 0.73 & 0.58 \\
StateAct (self-prompt+react) & \textbf{0.53} & 0.39 & \textbf{0.80} & \textbf{0.91} & 0.64 & 0.65 \\
StateAct (self-prompt+state+action) & 0.14 & \textbf{0.60} & 0.71 & 0.86 & 0.51 & 0.56 \\
StateAct (self-prompt+state+react) & 0.49 & 0.46 & 0.78 & 0.90 & \textbf{0.76} & \textbf{0.68} \\
\hline
\end{tabular}
\caption{135 Test Environments from Alfworld. Different Columns represent different models. 
In \textbf{Bold}: Best \& 2nd Best Solution per Model. Light Green Background: Best Solution per Model. Dark Green Background: Best Solution Overall. 
Decoding Strategy: Greedy (temperature=0). M=Mistral-Instruct-2501, Q=Qwen2.5; G=Gemma 2-Instruct.}
\label{tab:results_alfworld}
\end{table*}

\section{Additional results Webshop}
\label{appendix:additional_webshop}
For Webshop, we present the results for ReAct and StateAct. Similarly to Alfworld, we also present the results of StateAct in different forms, see Table \ref{tab:results_webshop}.

We can see that our method again outperform the base-agents. For most model we see big jumps in improvement using `state' in the Webshop environment, indicating that the environment benefits from additional structured prediction.

Interestingly in Webshop the results of Act-only are very strong for many models. A hypothesis might be that strong instruction tuning introduces these kind of reasoning techniques into the models directly without the need for explicit prompting from the user. Additionally, we can see that thoughts often harm performance on Webshop. A hypothesis for this is that Webshop already has a lot of textual content so verbose thoughts can be harmful by confusing the model.



\begin{table*}[h]
\centering
\begin{tabular}{|l|ccccc|c|}
\hline
\textbf{Agent Name} & \textbf{M-24B} & \textbf{Q-7B} & \textbf{Q-14B} & \textbf{Q-32B} & \textbf{G-27B} & \textbf{Average}\\
\hline
\textbf{Baselines}&\multicolumn{5}{l|}{\textbf{}} &\\
\hline
act-only (action) & 0.32 & 0.26 & 0.16 & 0.11 & 0.11 & 0.19 \\
react (thought + action) & 0.34 & 0.19 & 0.22 & 0.27 & 0.26 & 0.26 \\
\hline
\textbf{Our Methods}&\multicolumn{5}{l|}{\textbf{}} &\\
\hline
StateAct (state+action) & \textbf{0.38} & 0.25 & \textbf{0.37} & 0.29 & 0.26 & \textbf{0.31} \\
StateAct (self-prompt+action) & 0.00 & \textbf{0.27} & 0.36 & 0.23 & \textbf{0.35} & 0.24 \\
StateAct (state+react) & 0.23 & 0.14 & 0.25 & 0.26 & 0.04 & 0.18 \\
StateAct (self-prompt+react) & 0.24 & 0.14 & 0.15 & 0.25 & 0.14 & 0.18 \\
StateAct (self-prompt+state+action) & 0.37 & 0.26 & 0.36 & 0.30 & 0.15 & 0.29 \\
StateAct (self-prompt+state+react) & 0.35 & 0.12 & 0.33 & \textbf{0.32} & 0.29 & 0.28 \\
\hline
\end{tabular}
\caption{100 Test Environments from Webshop. Different columns represent different models. Average is the average of all models.
Decoding Strategy: Greedy (temperature=0). M=Mistral-Instruct-2501, Q=Qwen2.5; G=Gemma 2-Instruct.}
\label{tab:results_webshop}
\end{table*}

\section{Additional results for Textcraft}
We can see that in Textcraft similar to Alfworld. The results improve when thoughts are used. Interestingly, for Textcraft self-prompting does not yield the highest result. A hypothesis for is because the crafting recipe needs to followed very closesly for successful completion therefore additional reminders are not necessary.
\label{appendix:additional_textcraft}

\begin{table*}[h]
\centering
\begin{tabular}{|l|ccccc|c|}
\hline
\textbf{Agent Name} & \textbf{M-24B} & \textbf{Q-7B} & \textbf{Q-14B} & \textbf{Q-32B} & \textbf{G-27B} & \textbf{Average}\\
\hline
\textbf{Baselines}&\multicolumn{5}{l|}{\textbf{}} &\\
\hline
act-only (action) & 0.37 & 0.05 & 0.26 & \textbf{0.42} & 0.25 & 0.27 \\
react (thought + action) & 0.33 & 0.02 & 0.31 & 0.31 & 0.18 & 0.23 \\
\hline
\textbf{Our Methods}&\multicolumn{5}{l|}{\textbf{}} &\\
\hline
StateAct (state+action) & 0.34 & 0.06 & 0.09 & 0.36 & 0.25 & 0.22 \\
StateAct (self-prompt+action) & 0.42 & 0.01 & 0.25 & 0.41 & 0.23 & 0.26 \\
StateAct (state+react) & \textbf{0.45} & \textbf{0.12} & \textbf{0.38} & \textbf{0.42} & \textbf{0.34} & \textbf{0.34} \\
StateAct (self-prompt+react) & 0.41 & 0.09 & 0.35 & 0.37 & 0.29 & 0.30 \\
StateAct (self-prompt+state+action) & 0.29 & 0.01 & 0.16 & 0.35 & 0.23 & 0.21 \\
StateAct (self-prompt+state+react) & 0.40 & 0.04 & 0.37 & 0.40 & \textbf{0.34} & 0.31 \\
\hline
\end{tabular}
\caption{Performance across 100 Textcraft test environments with different models.
Decoding Strategy: Greedy (temperature=0). M=Mistral-Instruct-2501, Q=Qwen2.5; G=Gemma 2-Instruct.}
\label{tab:results_textcraft}
\end{table*}

\section{Use of AI}
We used coding assistants in small parts using continue.dev \footnote{\url{https://www.continue.dev/}, last accessed March 2025.} and claude-sonnet-3.5. Small use of ChatGPT was used for Latex advise and particular phrasing of parts of the text.